# Deep Reinforcement Learning for Decentralized Multi-Robot Exploration With Macro Actions

Aaron Hao Tan, *Student Member, IEEE*, Federico Pizarro Bejarano, *Student Member, IEEE*, Yuhan Zhu, Richard Ren, and Goldie Nejat, *Member, IEEE*

***Abstract*—Cooperative multi-robot teams need to be able to explore cluttered and unstructured environments while dealing with communication dropouts that prevent them from exchanging local information to maintain team coordination. Therefore, robots need to consider high-level teammate intentions during action selection. In this letter, we present the first Macro Action Decentralized Exploration Network (MADE-Net) using multi-agent deep reinforcement learning (DRL) to address the challenges of communication dropouts during multi-robot exploration in unseen, unstructured, and cluttered environments. Simulated robot team exploration experiments were conducted and compared against classical and DRL methods where MADE-Net outperformed all benchmark methods in terms of computation time, total travel distance, number of local interactions between robots, and exploration rate across various degrees of communication dropouts. A scalability study in 3D environments showed a decrease in exploration time with MADE-Net with increasing team and environment sizes. The experiments presented highlight the effectiveness and robustness of our method.**

## I. Introduction

MULTI-ROBOT exploration addresses the problem of a robot team exploring an unknown environment to obtain perception knowledge for spatial reasoning and awareness [1]. This is critical for cooperative robot applications ranging from warehouse automation [2], to search and rescue (SAR) [3]. To date, multi-robot coordination methods can be categorized as centralized or decentralized. Centralized systems can achieve optimal performance in terms of overall mission time and travel distance; however, they suffer from single-point failures, poor scalability, and their reliance on communication to exchange sensory information for decision-making [4]. Therefore, these systems are not well-suited in environments with unreliable communication. Alternatively, decentralized systems coordinate through local communication with neighboring robots [5]. However, traditional architectures require strong domain knowledge for handcrafted heuristics to achieve the desired cooperation behaviors [2]; thus, are limited to simple and specific tasks and environments [6]. Additionally, robot coordination is dependent on locally exchanged information, which results in overall degraded team performance during communication dropouts, as robots only account for local goal attributes (i.e., distance and time to travel) without consideration of teammates' intentions [7].

Recently, deep reinforcement learning (DRL) methods have been used to solve challenging domains by learning directly from the experiences collected by the robots within the task environment [8]. These methods include our previous DRL work for both robot navigation [9], [10], and exploration [11] in unknown cluttered environments. However, only a handful have incorporated DRL to solve multi-robot exploration problems [5], [6], [12], [13], [14]. These methods have mainly employed centralized systems that depend on constant information sharing between the robots.

Macro actions are temporally extended actions consisting of a series of primitive actions that last a single timestep [15]. As a result, macro actions describe high-level robot intentions (i.e., expected goals to visit) instead of low-level robot actions (i.e., directional movements), which enable asynchronous decision making that directly reasons about goal selection and teammates' expected goals. This is advantageous for avoiding redundant coverage during multi-robot exploration with communication dropouts. However, to-date, DRL-based macro action planners have only been applied for material handling and warehouse delivery [16], [17] problems in known environments with a priori knowledge of task sequences and locations; and have not yet been applied to multi-robot exploration in unseen environments.

In this paper, we introduce the first DRL-based decentralized multi-robot exploration approach that implicitly accounts for teammate intentions during communication dropouts in unknown environments. Namely, our unique contributions are: 1) the development of a new macro action based decentralized exploration planner, called MADE-Net, that introduces the utilization of a map feature extractor (MFE) network in order to uniquely consider the spatial context of high-level robot tasks within their environments. This enables MADE-Net to generalize to unseen environments where task locations and completion sequences are unknown and can vary; and 2) the introduction of a communication dropout technique using a communciation success probability (CSP) to determine whether local teammate intentions are exchanged, in order to learn communication invariant features for implicit consideration of teammate intentions during decentralized execution.

Manuscript received 8 September 2022; accepted 3 November 2022. Date of publication 24 November 2022; date of current version 8 December 2022. This letter was recommended for publication by Associate Editor X. Yu and Editor M. A. Hsieh upon evaluation of the reviewers' comments. This work was supported in part by the Natural Sciences and Engineering Research Council of Canada (NSERC), an NSERC Collaborative Research and Development (CRD) grant, and in part by the Canada Research Chairs program (CRC). *(Corresponding author: Aaron Hao Tan.)*

The authors are with the Autonomous Systems and Biomechatronics Laboratory (ASBLab), Department of Mechanical and Industrial Engineering, University of Toronto, Toronto, ON M5S 3G8, Canada (e-mail: aaronhao.tan@utoronto.ca; federico.pizarrobejarano@mail.utoronto.ca; yuhanelle.zhu@mail.utoronto.ca; richa.ren@mail.utoronto.ca; nejat@mie.utoronto.ca).

## II. Related Works

Existing multi-robot exploration methods can be categorized as: 1) classical [7], [18], [19], [20], [21], [22], [23], [24], [25], [26], [27], [28], [29] or 2) deep reinforcement learning [5], [6], [12], [13], [14].

### A. Classical Methods for Multi-Robot Exploration

Classical decentralized methods use handcrafted heuristics for robot goal allocation [1]. They can be either utility-based [18], [22], [23], [25], [26], [27], [30], market-based [19], [20], [21], [24], [28], [29], or planning-based [7].

Utility-based and market-based methods utilize a utility function to maximize the robot information gain [22], [23], or an auction system to select goal bids based on estimated profit [28], [29], respectively. In both methods, the value of a goal is estimated by the expected information gain [1], [25], or place semantics [18], [26], while the costs include the travel distance [25], [27], [30], time to goal [19], [20], [21], [24], and the distance to teammates [25], [27].

In [7], a planning-based method was proposed using Decentralized Markov Decision Process (Dec-MDP) and solved via value iteration online. During communication dropouts, the probability of a robot teammate in a particular state, given its last known state and the timesteps elapsed since the last observation, was incorporated into the state value estimation.

### B. DRL Methods for Multi-Robot Exploration

Using DRL, robots can learn complex exploration strategies through repeated interactions with their environments [31]. Existing approaches utilize either centralized [5], [6], [12], or decentralized [13], [14] approaches during execution.

In [5], a CommNet-Explore model was used to learn multi-robot cooperation behaviors in a dynamic 2D grid world. The model consists of a communication channel where robots shared their observations of local cells and trajectories for coordination. This was extended in [6] with an attention mechanism to enable communication with only specific teammates based on the relevancy of the messages. In [12], global map images that contained robot positions, obstacles, and explored/unexplored regions, were used as inputs to a Proximal Policy Optimization (PPO) model to generate joint primitive directional actions for the robot team.

In [13], [14], robots used centralized training for decentralized execution (CTDE) to learn primitive exploration policies that only required locally exchanged metric and topological maps in structured open space environments. These methods used multi-agent deep deterministic policy gradient (MADDPG) [13], and multi-agent PPO [14], for coordination in unknown environments.

### C. Summary of Limitations

In summary, classical methods utilize handcrafted heuristics that require extensive manual tuning of utility and cost functions to achieve the desired cooperative behaviors [11]. DRL methods remove the need for handcrafted heuristics and learn cooperation strategies directly from robot experience [6]; however, they are either: 1) centralized and require constant communication for coordination [5], [6], [12], or 2) planning occurs at the primitive-level without consideration for high-level teammate intentions which can result in redundant area coverage during communication dropout [13], [14]. Namely, primitive-based methods focus on short-term coverage rewards (i.e., area explored per primitive action), which do not directly transfer to robots selecting complementary goals to visit (i.e., area explored per macro action).

To the authors' knowledge, DRL-based macro action planners have only been proposed for multi-robot material handling and warehouse delivery tasks [16], [17], where robots have full communication with each other, and communication dropouts are not considered during execution. Furthermore, these environments are a priori known with pre-defined task locations, therefore, a fixed static task completion sequence is executed. To address these limitations, we introduce MADE-Net, the first macro action based multi-robot DRL approach for decentralized exploration that uniquely accounts for high-level teammate intentions during communication dropouts. We are the first to introduce a macro action planner that can: 1) generalize to various unseen, unstructured and cluttered environments by utilizing a MFE network to learn spatial invariant features, and 2) implicitly account for teammate intentions during decentralized exploration by utilizing a communication dropout technique during centralized training to learn communication invariant features.

## III. The Multi-Robot Exploration Methodology

In this section, we define the multi-robot exploration problem and present our novel MADE-Net architecture.

### A. Problem Definition

The decentralized multi-robot exploration problem consists of a team of mobile robots, $I = \{I_1, ..I_n\}$, cooperatively exploring an unknown environment while generating a global map, $\mathcal{M}$. The environment is represented by a grid map, whose configuration is cluttered and unstructured (e.g., containing irregular-shaped obstacles). Each robot is equipped with perception sensors with a fixed sensing range, $d_s$. During deployment, robots can exchange their local map $m_i$, local position, $x_i \in X$, and exploration goal positions, $g_i \in X$, if within: 1) communication range (defined by the sensing range, $d_s$), and 2) line of sight (LoS), not obstructed by obstacles from each other. The objective is to maximize the combined area explored, $|E|$, over the joint distance traveled, $D$, for each discrete timestep, $t$, during exploration:

$$\max \left[ \sum_{j=1}^{h} D_j^{-1} \, |E|_j \right], \tag{1}$$

where $h$ denotes the total number of timesteps in an episode.

### B. MacDec-POMDP Model for Multi-Robot Exploration

We model the decentralized multi-robot exploration problem as a Macro Action Decentralized Partially Observable Markov Decision Process (MacDec-POMDP) [32]. A MacDec-POMDP is defined by the tuple $\langle I, S, \{M_i\}, \{A_i\}, T, R, \{Z_i\}, \{\Omega_{\mathrm{I}}\}, \zeta_i, O, h\rangle$. Herein, $S$ denotes the set of environment states. Each exploration action is represented as a macro action which requires a sequence of primitive actions, $a_i \in A_i$, over a duration of timesteps to complete. The set of available macro actions for robot $i$ is denoted as, $M_i \in M$.

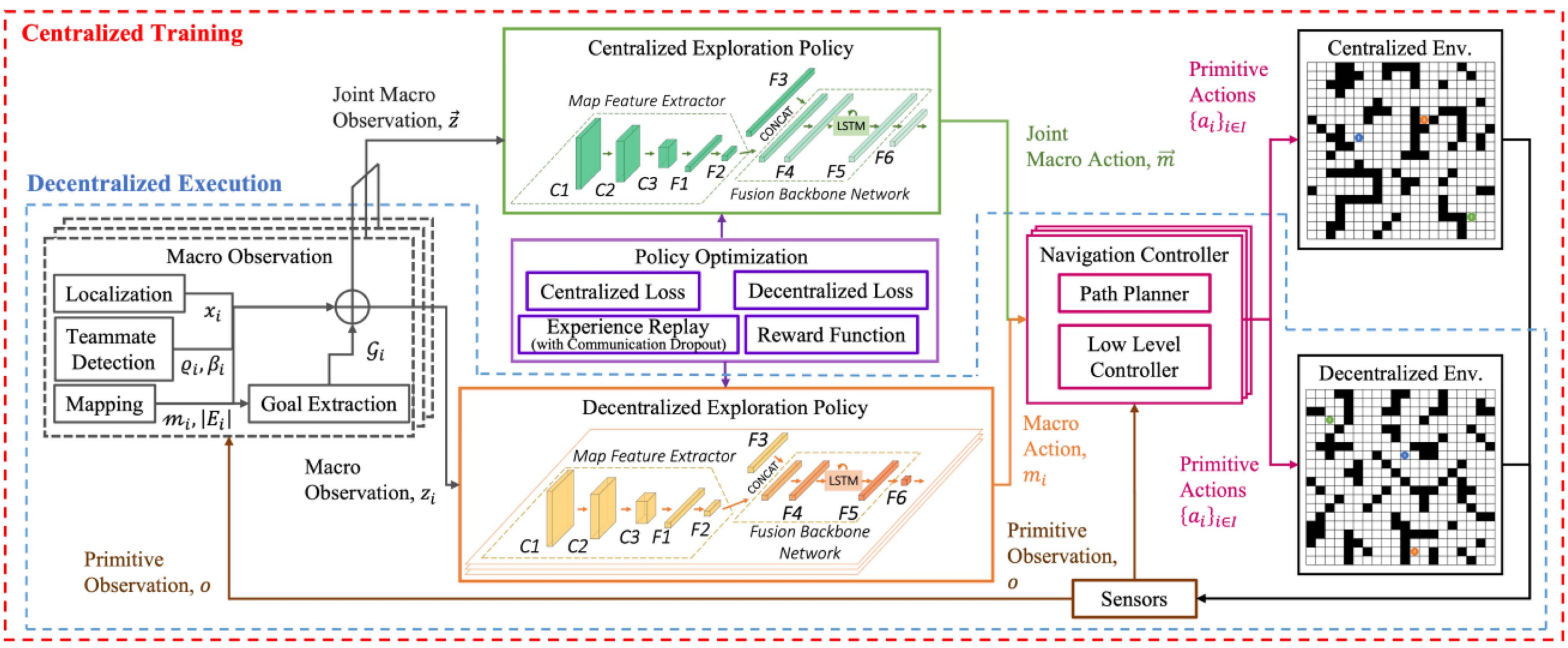

Fig. 1. MADE-Net architecture with centralized training and decentralized execution. The grid world environment (right) includes obstacles (black), free space (white) and robots (blue, orange, green). Convolution, concatenation and fully connected layers are denoted by C, CONCAT, F, respectively.

A reward is determined by the reward function, $R$, based on $a_i$, taken in $s$ at $t$. Macro observations, $z_i \in \zeta_i$, are high-level representations of the environment (i.e., robot and teammate positions, maps), while primitive observations, $o_i \in \Omega_i$, are sensory depth data used to extract cell occupancy for obstacle and teammate detection. $o_i$ and $z_i$ are generated by probability functions $O_i(o_i,\ a_i,\ s)$, and $Z_i(z_i, m_i, s)$, respectively. Robots receive $z_i$ at each timestep where macro actions are terminated $\vec{\tau}$, while $o_i$ is received at each $t$. The transition function is defined by $T(s', \vec{\tau},\ s,\ \vec{m})\ = \Pr\left(s', \vec{\tau} \mid s, \vec{m}\right)$, where $\vec{m}$ is the joint macro action. The goal is to find a set of decentralized policies, $\Psi_{\mathrm{I}}$, where the value of the combined joint policy, $\Psi\ = \{\Psi_{\mathrm{i}}\}_{i\in I}$, is optimized to achieve maximum expected team rewards over the horizon, $h$, of an episode. $s_o$ is the initial state, and $\gamma$ is a discount factor for future rewards, $r$ [17]:

$$\Psi^{*} = \arg\max_{\Psi} \mathbb{E}\left[\sum_{t=0}^{h-1} \gamma^{t}\ r\left(s\left(t\right),\ \vec{a}\left(t\right)\right) | s_0, \Psi\right]. \tag{2}$$

## C. Multi-Robot Exploration Architecture

The novel DRL macro action based multi-robot exploration architecture proposed in this work is presented in Fig. 1, which consists of both the *Centralized Training* and *Decentralized Execution* sub-systems. The objective is to learn robot team behaviors and intentions with the *Centralized Training* sub-system while executing team exploration with local perception using the *Decentralized Execution* sub-system. The *Policy Optimization* module updates both the *Centralized and Decentralized Exploration Policy* during training where macro actions are executed simultaneously via the *Navigation Controller* module using parallel environments. The following sections discuss the main modules of the architecture in detail.

*1) Macro Observations:* The proposed *Macro Observation* module features four submodules; namely, *Localization*, *Teammate Detection*, *Mapping* and *Goal Extraction*, to process the $o_i$ obtained from the robot *Sensors* into $z_i$. Specifically, *Localization* determines the robot's global position, $x_i \in X$, with respect to a world coordinate frame. *Teammate Detection* outputs a Boolean value, $\varrho_i$, to indicate if a teammate is observed within $d_s$. During exploration, the following information is exchanged between robots within $d_s$, $\beta_i = [\{g_j\}_{j\in I'}^{t-1}, \{x_j\}_{j\in I'}^{t}, \{g_j\}_{j\in I'}^{t}, \{m_j\}_{j\in I'}^{t}]$, where $I'$ denotes the list of teammates. This exchange includes the observed teammate's previous exploration goal, $\{g_j\}_{j\in I'}^{t-1}$, current position, $\{x_j\}_{j\in I'}^{t}$, current exploration goal, $\{g_j\}_{j\in I'}^{t}$, and local map, $\{m_j\}_{j\in I'}^{t}$. For unobserved robots in the team at $t$, $\beta_i$ includes the last observation of these robots, and thus remains unchanged. *Mapping* performs both mapping and map merging. Each robot updates its local map, $m_i \in \mathcal{M}$, and the area explored, $|E_i|$, using its sensors. A map describes a four channel image consisting of binary feature maps for the explored space, obstacles, observed robot positions, and goal candidates. Local maps are merged when robots are within $d_s$ to update the explored regions. A global map, $\mathcal{M}$, is generated for centralized training, by merging all $m_i$ at each $t$.

The *Goal Extraction* submodule uses $m_i$, and $x_i$, to cluster frontier locations into a set of goal candidates, $\mathcal{G}_i$. The number of goal candidates available to each robot, $|\mathcal{G}_i|$, is dependent on the environment size $|\mathrm{X}|$ team size $|\mathrm{I}|$ and sensing range $d_s$, and is described by a logarithmic relationship:

$$4\log\left(\frac{\sqrt{|\mathrm{X}|}}{2\mathrm{d_s}+1}\right) + |\mathrm{I}|\,, \text{where}\ \left(2d_s+1\right) \leq \sqrt{|X|}. \tag{3}$$

Eq. (3) ensures the minimum of $|\mathcal{G}_i|$ is equal to $|\mathrm{I}|$ to provide each robot with a unique location to explore. Increasing $|\mathrm{X}|$ and $|\mathrm{I}|$ result in increasing $|\mathcal{G}_i|$, where the Euclidean distance between each goal candidate, $\mathcal{g}_n \in \mathcal{G}_i$, is maximized to provide a set of spatially distributed goals. Conversely, an increasing $d_s$, decreases $|\mathcal{G}_i|$, to reduce overlap of sensor coverage within the environment. During exploration, $\mathcal{G}_i$ is updated to account for new exploration goals based on the explored map, $m_{\mathrm{I}}$, at each $\vec{\tau}$.

The outputs of the *Macro Observation* submodules are combined to create the individual macro observation, $z_i =$

$\langle x_i, \varrho_i, \beta_i, m_i, |E_i|, \mathcal{G}_i \rangle$, and the joint macro observation, $\vec{z} = \langle \{x_i\}_{i\in I}, \{\varrho_i\}_{i\in I}, \{\beta_{\mathrm{i}}\}_{i\in I}, \mathcal{M}, |E|, \{\mathcal{G}_i\}_{i\in I} \rangle$. $\vec{z}$ and $z_i$ are sent to the *Centralized Exploration Policy* and *Decentralized Exploration Policy*, respectively, for macro action selection.

*2) Centralized Exploration Policy (CEP):* The *CEP* is approximated using Deep Double Recurrent Q Networks (DDRQN), which includes an estimation, $Q$, and a target network, $Q'$, in its update rule [33]:

$$Q(\delta, a) \leftarrow Q(\delta, a) + \alpha \left[ R + \gamma Q'(\delta', a^*) - Q(\delta', a) \right], \quad (4)$$

where $a^* = \mathrm{argmax} Q(\delta', a')$, $\alpha$ is the learning rate, $\gamma$ is the discount factor, and $\delta$ denotes the observation history.

To generalize to unseen environments without the availability of pre-defined spatial task locations and execution sequences, we introduce the first macro action planner that learns the spatial context of high-level robot tasks and environment configurations. Namely, MADE-Net utilizes a novel MFE network with convolutional layers to extract spatial invariant features from $\mathcal{M}$ for decentralized planning. The MFE includes three convolutional layers (C1-C3) followed by F1 and F2, which are fully connected layers (FCLs). The remaining observations, $\vec{\mathrm{z}}/\{\mathcal{M}\}$, are encoded by the FCL, F3. A concatenation layer, CONCAT, combines the multi-modal observations from F3 and the MFE. The combined features are passed through a FCL, F4, and a Long Short Term Memory (LSTM) layer to learn temporal features from past observations and actions to account for the spatial correlation between robot teammate intentions and the environment. F5-6 are used to learn a non-linear combination of temporal features, whose outputs are the *Q* values for each permutation of joint macro actions.

The joint macro action space of the team is represented by $M = \{M_i\}_{i\in I}$, where $M_i = [m_1, m_2, m_3, m_4]_i$ denotes the set of macro actions for each robot $i$. Each macro action in $M_i$ navigates the robot to the corresponding exploration goal candidate in $\mathcal{G}$. Thus, $m_1$, describes a tuple, $\langle \beta^m, \mathcal{I}^m, \pi^m \rangle$, where the initial condition, $\mathcal{I}^m$, consists of all free space within the environment. The termination condition, $\beta^m$, happens when the robot arrives at the intended goal location, or when a teammate is detected during the execution of the macro action. The low-level policy, $\pi^m$, is used in the *Navigation Controller* module, to generate primitive actions based on primitive observations. The selected joint macro action, $\vec{m}$, is sent to the *Navigation Controller* for execution in the centralized environment.

*3) Decentralized Exploration Policy (DEP):* The *DEP* also uses DDRQN and shares the same architecture as *CEP*, but differs in layer dimensions for F3, CONCAT, F4-6, to account for the smaller observation and action space. The $m_i$ selected by the *DEP* is executed by the *Navigation Controller* in the decentralized environment.

*4) Policy Optimization:* The objective of the *Policy Optimization* module is to train both the *CEP* and *DEP* simultaneously. This is completed with a *Centralized* and *Decentralized Loss* function, *Reward Function*, and an *Experience Replay* submodule.

*Centralized and Decentralized Loss Function:* The *Centralized Loss* function, $\mathcal{L}(\phi)$, is defined as [17]:

$$\mathbb{E}\left[\left(\begin{matrix} r + \gamma Q'_\phi \left(\delta', \underset{\vec{m}'}{\arg\max}\ Q_\phi\left(\delta', \vec{m}' | \vec{m}^{\mathrm{ud}}\right)\right) \\ -Q_\phi(\delta, \vec{m}) \end{matrix}\right)^2\right], \quad (5)$$

where $Q_\phi$ and $Q'_\phi$ are the estimation and target network of *CEP*, respectively. $\mathrm{r}$ denotes the team reward. $\delta$ and $\delta'$ are the current and next joint macro observation history while $\vec{m}$ and $\vec{m}'$ are the current and next joint macro action. $\vec{m}^{\mathrm{ud}}$ denotes the status of macro actions from robot's whose macro actions have not terminated. The *Decentralized Loss* function, $\mathcal{L}(\theta_i)$, is defined as [17]:

$$\mathbb{E}\left[\left(\begin{matrix} \mathrm{r_i} + \gamma Q'_{\theta_i} \left(\delta'_i, \left[\underset{\vec{m}'}{\arg\max}\ Q_\phi\left(\delta', \vec{m}' | \vec{m}^{\mathrm{ud}}\right)\right]_i\right) \\ -Q_{\theta_i}(\delta_i, m_i) \end{matrix}\right)^2\right]_{i\in I}, \quad (6)$$

where $Q_{\theta_i}$ and $Q'_{\theta_i}$ are the estimation and target network of *DEP* for robot $i$. $\mathrm{r}_i$ is the robot's reward and $\delta_i$ and $\delta'_i$ are the current and next individual macro observation history. The *Decentralized Loss* function uses $Q_\phi$ from *CEP* to select $\vec{m}'$, where the individual macro action, $m_i \in \vec{m}'$, for robot $i$ becomes the target action. As a result, the *Decentralized Loss* function incorporates action selection from the *CEP*; which enables the *DEP* to account for their teammates' intentions using centralized information during training, without direct information exchange during execution.

*Reward Function:* Rewards are designed to encourage robots to maximize the joint area explored while minimizing local interactions, distance traveled, and time to completion. Local interactions are defined when two or more robots are within $d_s$ sensing range. Rewards for individual robots $(\mathrm{r_i})$, and the team $(\mathrm{r})$ are:

$$\mathrm{r_i} = \begin{cases} -\left(c_i^t \cap E_i^t\right), & \text{each } t \\ \left(c_i^t \cap U_i^t\right), & \text{each } t \\ -(1/5 * e^\rho)^{0.5}, & \rho = [2, 7] \\ -15, & \rho > 7 \end{cases}, \quad (7)$$

$$\mathrm{r} = \begin{cases} -\left(C_t \cap E_t\right), & \text{each } t \\ \left(C_t \cap U_t\right), & \text{each } t \\ -1, & \text{each } t \\ +100, & s_{complete} \end{cases}, \quad (8)$$

$\mathrm{c_i^t}$ denotes the set of cells within $d_s$, at $t$ for robot $i$. Each robot receives a negative reward for every observed cell in the explored set $E_i^t$ and a positive reward in the unexplored set $U_i^t$. $\rho$ denotes a count of consecutive local interactions where an exponentially increasing negative reward is given to all robots within $d_s$, between $\rho$ of 2 and 7, up to a maximum of $-15$. Local interactions enable robots to exchange information for coordination $(\rho < 2)$; while prolonged interactions $(\rho > 7)$ can lead to conflicts such as collisions and/or deadlocks [7]. $C_t$ denotes the set of cells observed by the entire team at $t$, while $E_t$ and $U_t$ are the set of jointly explored and unexplored cells. The team receives a positive/negative reward for every cell that is previously unobserved/observed, respectively. A negative reward is given at each $t$ to minimize the total exploration timesteps, while a reward of $+100$ is assigned at terminal state, $s_{complete}$, to encourage cooperation.

*Experience Replay and Parallel Environments:* Two environments with separate teams of robots are used to train the *CEP* and *DEP* in parallel [17]. In both environments robots collect experiences defined by $\langle z_i, m_i, z'_i, \mathrm{r_i} \rangle$, and $\langle \vec{z}, \vec{m}, \vec{z}', \mathrm{r} \rangle$, in the decentralized and centralized environments, respectively, and store them in the *Experience Replay* buffer at each $t$. To address the challenge of communication dropouts, we introduce a CSP which is sampled from a uniform distribution, $p \sim U(0, 1)$, during centralized training within the decentralized environment. The CSP is used to determine the probability of robots within $d_s$ updating their observation vectors with $\beta_i$ to learn communication invariant features from decentralize robot experiences with missing teammate information. Thus, the *DEP* learns to implicitly account for teammate intentions in an end-to-end manner by using communication invariant features and joint macro actions from the *CEP,* Eq. (5). $\epsilon -$ greedy with a linear decay rate is incorporated to expose the *DEP* to cooperative robot actions that have a higher likelihood of occurring in the *CEP*'s environment during training. A mini-batch of experiences is uniformly sampled from the *Experience Replay* during policy updates [34].

*5) Navigation Controller:* The *Navigation Controller* module executes primitive robot navigation actions to complete the selected macro action. However, multiple robots navigating within a shared space must consider each other's actions to avoid local conflicts [35]. This problem is addressed by: 1) negatively rewarding local interactions between robots for prolonged periods, and 2) incorporating spatial-temporal A$^*$ [36] to consider the spatial configuration of the environment as well as the temporal aspects of nearby robot trajectories to resolve deadlock scenarios through prioritized planning. The path is executed by the *Low-Level Controller* which generates primitive directional actions based on the robot's drive system and the cell occupancy feedback from the *Sensor* module until the macro action termination condition is met.

## IV. Training

MADE-Net was trained in parallel $20 \times 20$ m$^2$ grid world environments with 1 m $\times$ 1 m cells, using two separate robot teams controlled by the *CEP* and *DEP,* respectively. At the start of each episode, robot teams of three are spawned randomly in environments with obstacle densities between 40% to 60%, to incorporate a variety of cluttered scenes. Each robot has a $d_s$ of four cells based on its LoS, while an episode ends when the environment is fully mapped by the team.

The MFE network's kernel size, stride, and output channel for C1, C2, and C3 are (4, 2, 8), (3, 2, 16), and (2, 2, 16), respectively. F1 and F2 include 32 and 10 neurons each. The *CEP* and *DEP* has 128, and 64 neurons, respectively, in layers F3, CONCAT, F4-5. Both policies' LSTM layer has a hidden state size of 64. Since there are four macro actions available per robot, F6 in the *CEP* and *DEP* includes 64 and 4 neurons, respectively. Each layer except the output layer utilizes the leaky rectified linear units activation function.

The training was conducted on an AMD Ryzen Threadripper 3960X, 128 Gb RAM, for over 350 hours. A discount factor $\gamma$ of 1 and an experience batch size of 16 were used. Fig. 2 presents the team reward and the number of steps to complete exploration for each episode with an exponential weighted moving average of 0.9 during training. MADE-Net was able to converge within 250,000 episodes.

## V. Simulated Experiments

The simulated experiments include: 1) a comparison study to evaluate the performance of MADE-Net in 2D grid world environments with varying CSPs, and 2) a scalability study in 3D environments with varying team and environment sizes.

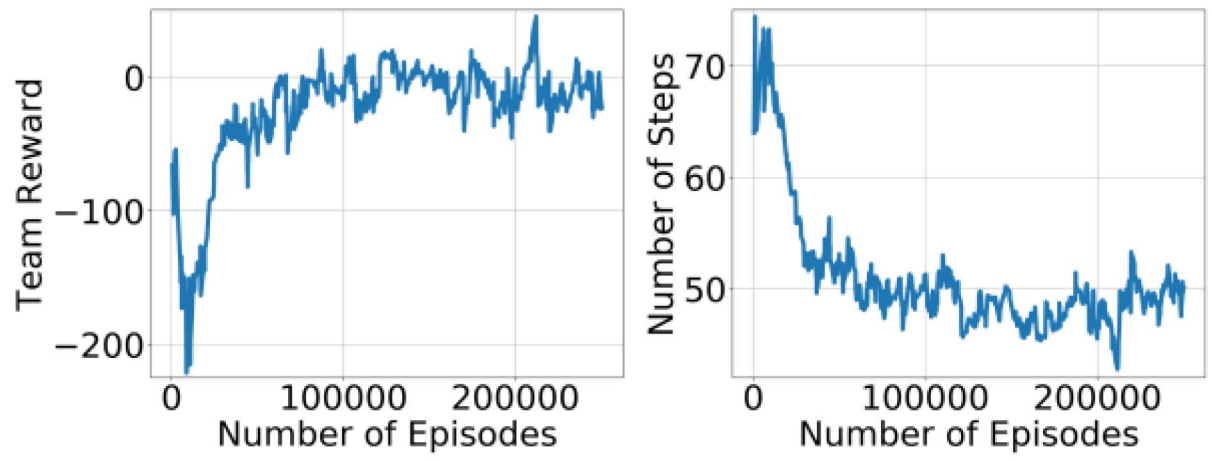

Fig. 2. Team reward (left), and the number of steps (right) during training with an exponential weighted moving average factor of 0.9.

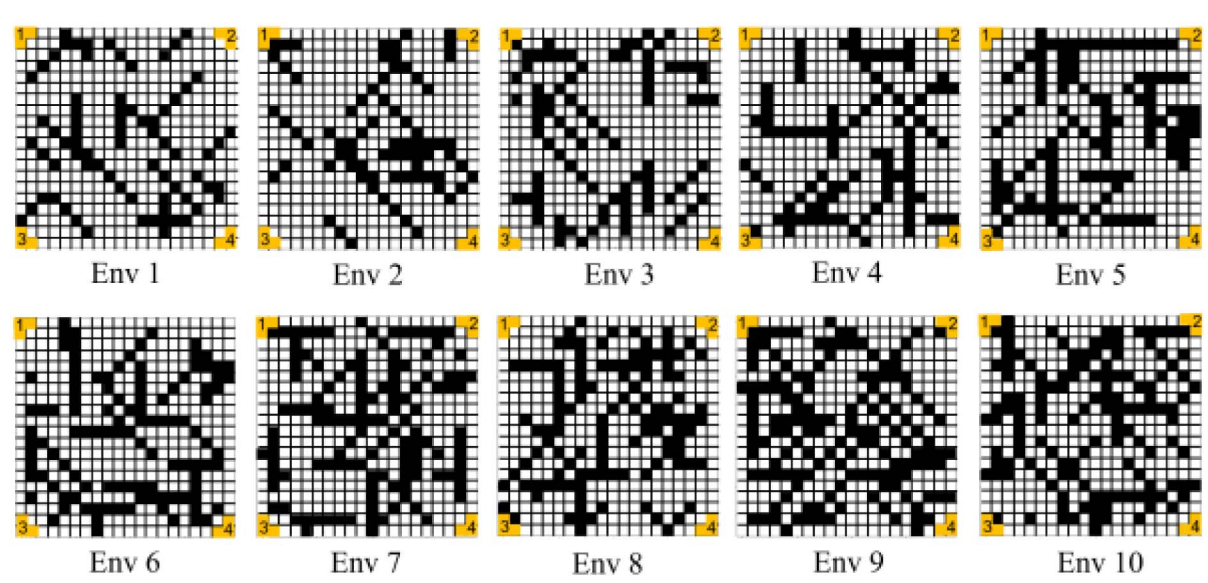

Fig. 3. 10 randomly generated $20 \times 20$ grid world environments with increasing clutter. Obstacles are black, and free spaces are white. Yellow cells represent the four starting team positions for each trial.

### A. Comparison Study

We evaluate MADE-Net's performance against classical and DRL exploration methods with respect to: 1) computation time (CT), 2), total number of steps (TNS), 3) total travel distance (TTD), 4) number of local interactions (NOLI), 5), objective function value (OFV), Eq. (1), and 6) exploration rate.

*1) Cluttered Environment:* Ten $20 \times 20$ unseen grid environments were randomly generated with increasing degrees of clutter to investigate spatial distribution within robot coordination, Fig. 3.

*2) Mobile Robots:* Teams of three robots were deployed where each cell within each robot's $d_s$ had a 10% probability of not being observed. Each robot also had a movement success probability of 90% in arriving at its desired cell to account for motion uncertainty.

*3) Communication Dropout:* Communication dropouts are represented by the failure of robots within $d_s$ to exchange information for a duration of 7 timesteps. In these experiments 0%, 50%, 80%, and 100% CSP were considered. During communication dropout, robots are still able to observe teammate positions.

*4) Procedure:* Four trials were conducted with four different initial team positions at the map corners shown in Fig. 3, for a total of 240 trials. A trial ends when the team has finished exploring the environment or a maximum timestep of 300 is reached. All 240 trials are repeated for each CSP for a total of 960 trials.

*5) Comparison Methods:* MADE-Net was compared against 3 classical primitive methods and 2 DRL methods: one primitive and one macro.

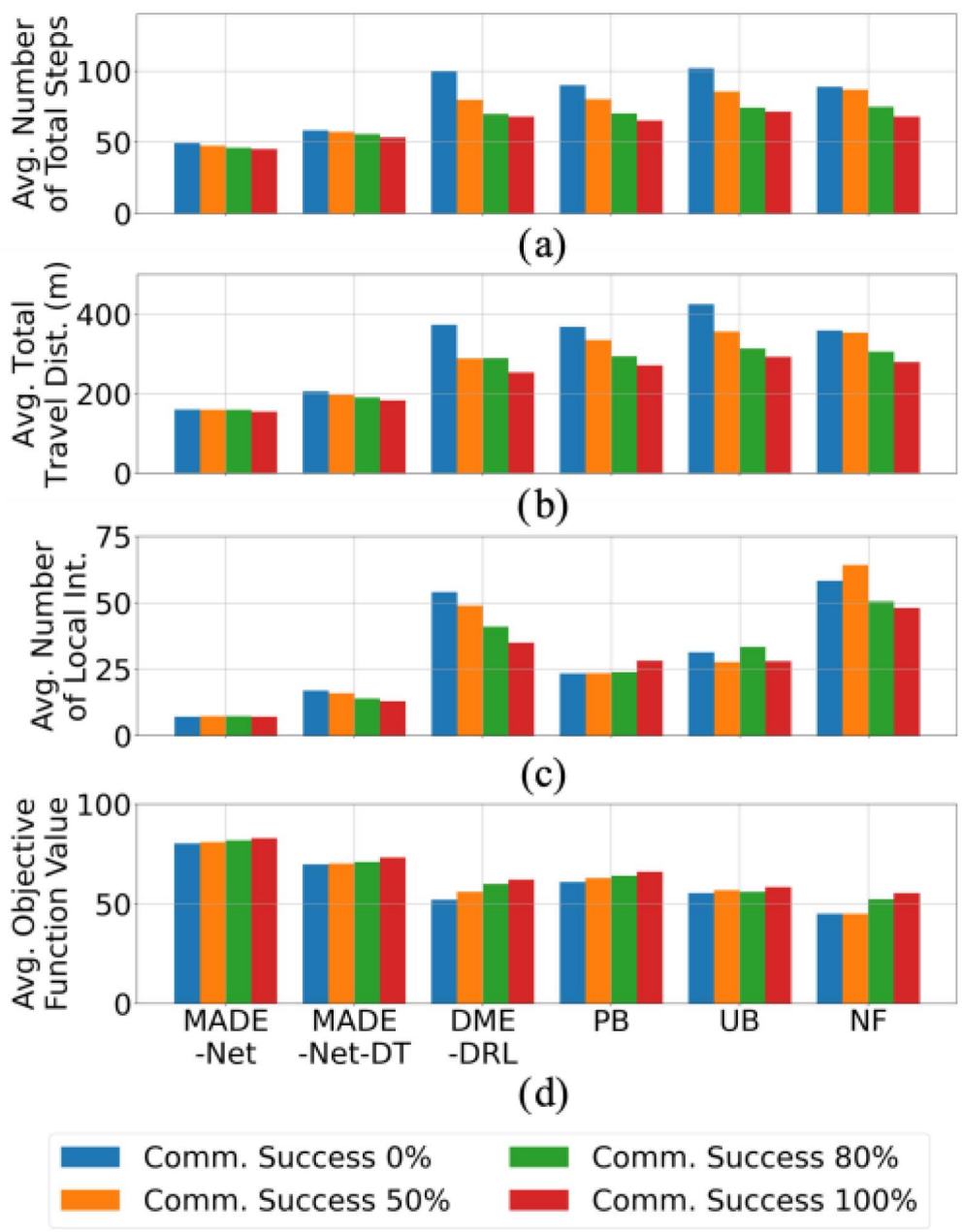

Fig. 4. Averages across all trials across communication success probabilities of 0%, 50%, 80% and 100% for: (a) total number of steps, (b) total travel distance, (c) number of local interactions, and (d) objective function values.

**Nearest Frontier Approach (NF):** The NF method [37] used a naïve strategy to visit the nearest frontier location from the robot's current position. Coordination is achieved through sharing local maps with neighboring robots.

**Utility-Based Approach (UB):** The UB method [25] utilized a utility function to incorporate information gain, distance cost, and a coordination factor to encourage spatial distribution during exploration.

**Planning-Based Approach (PB):** The PB method used a Dec-MDP for online planning [7] to directly account for communication dropout by predicting teammate states based on the last observed position and the timesteps elapsed. PB was adapted here to make use of square grids.

**Primitive Action DRL Approach (DME-DRL):** The DME-DRL [13] method used CTDE and MADDPG to learn robot coordination with primitive directional actions. Local information (teammate positions, maps) was exchanged when robots were within $d_s$.

**Macro Action DRL Approach (MADE-Net-DT):** MADE-Net-DT is a variant of MADE-Net with only decentralized training to investigate any performance difference in the learned policy.

*6) Results:* It took approximately 44s to complete 40 trials for MADE-Net and MADE-Net-DT, 73s for DME-DRL, 67s for NF, 116s for UB, and 2,040s for PB. PB had the longest CT as it required online value iteration, while all DRL-based methods only required a forward network pass for action selection.

The results for all methods across the four CSPs with respect to the average TNS, TTD, NOLI, and OFV, are shown in Fig. 4. In general, for 100% CSP, MADE-Net had the lowest average TNS (45), TTD (154m), NOLI (7), and highest average OFV (82) compared to the other methods, showing its ability to achieve better spatial distribution. Conversely, the PB and UB methods had a higher TNS (65 and 72) and TTD (271 m and 293 m) compared to MADE-Net as they prioritize high-reward states/goals early during the exploration, resulting in redundant coverage to visit previously unexplored cells. The NF method did not have a mechanism to distribute the robot team; therefore, robots stayed close together during exploration as evident by an average NOLI of 48 per trial. Using the Friedman Test, a statistically significant difference $(p < 0.0001)$ was found for each metric across all CSPs. Post-hoc analysis using Wilcoxon Signed-rank tests with Bonferroni correction of $\alpha_{revised} = 0.001$, showed a statistical difference $(p < 0.001)$ between MADE-Net and each classical method.

For the DRL methods, both DME-DRL and MADE-Net-DT yielded a higher average TNS (68 and 53), TTD (268 m and 183 m), and NOLI (35 and 13), and lower OFV (62 and 73) compared to MADE-Net. DME-DRL's lower performance is due to it only using locally exchanged maps to plan in the primitive action space, without considering robot intentions. Its reward function also does not explicitly punish local robot interactions, resulting in reduced spatial distribution during exploration. Furthermore, nearby robots received similar initial observations, which led to exploration of overlapping regions. MADE-Net-DT did not degrade as much as the other methods did compared to MADE-Net in terms of the TNS, and OFV. This is due to MADE-Net-DT also using macro actions to plan directly in the exploration goal space similar to MADE-Net. A statistically significant difference $(p < 0.001)$ was determined between MADE-Net, and both DME-DRL and MADE-Net-DT for all performance metrics across all CSPs using post-hoc Wilcoxon Signed-rank tests with Bonferroni correction of $\alpha_{revised} = 0.001$.

As the CSP decreases from 100% to 0%, MADE-Net maintained better performance compared to the benchmark methods. In particular, MADE-Net was able to outperform MADE-Net-DT as it was able to implicitly account for teammate intentions during communication dropouts, by using communication invariant features and centralized experiences during training. This is evident as the trend in the variances for average TNS, TTD, NOLI and OFV across all CSPs suggests that the performance of MADE-Net-DT was more affected by communication dropouts compared to MADE-Net. Similarly, the classical methods experienced increasing TNS and TTD with decreasing CSP as: 1) NF and UB methods required explicit information exchange for coordination, and 2) the cluttered environment forced robots to suddenly change their intentions during exploration; thereby, making it difficult for the PB method to effectively account for teammate intentions.

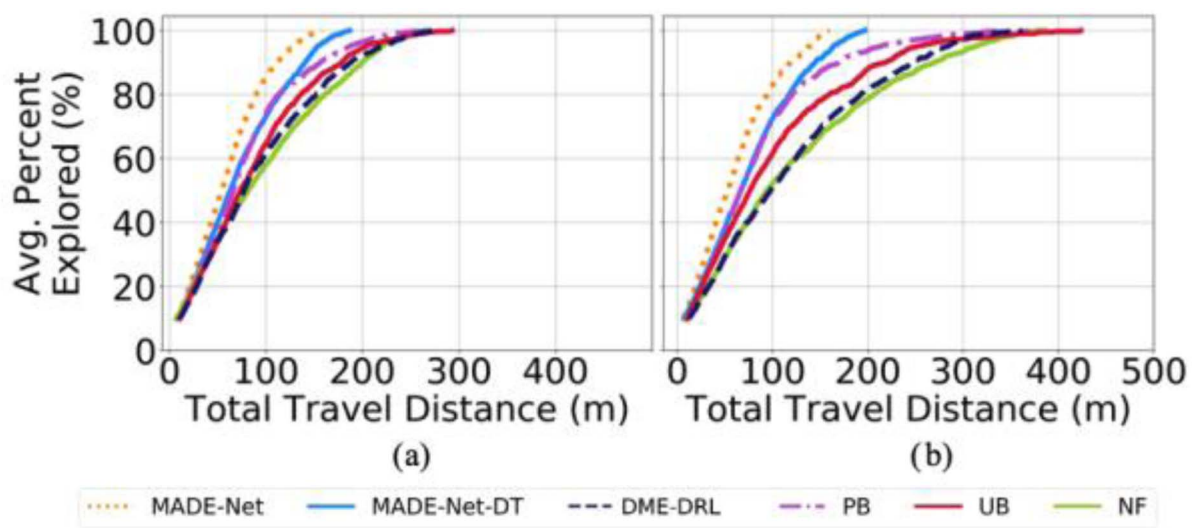

Fig. 5. Average percent exploration across all trials for (a) 100% communication success, and (b) 0% communication success.

Fig. 5 presents the average percent exploration of the environments with respect to the distance traveled across all trials for 100% and 0% CSP. At 100% CSP, primitive planners degrade the most after 80% explored while MADE-Net achieved a lower distance traveled throughout all percentages explored, as MADE-Net was trained to prioritize both distances traveled and coverage. This is advantageous for time sensitive exploration tasks such as SAR missions, where it is critical to cover a large area early in the exploration [38]. When considering 0% CSP, MADE-Net experienced minimal degradation in terms of travel distance compared to all benchmark methods, further demonstrating its robustness against communication dropout.

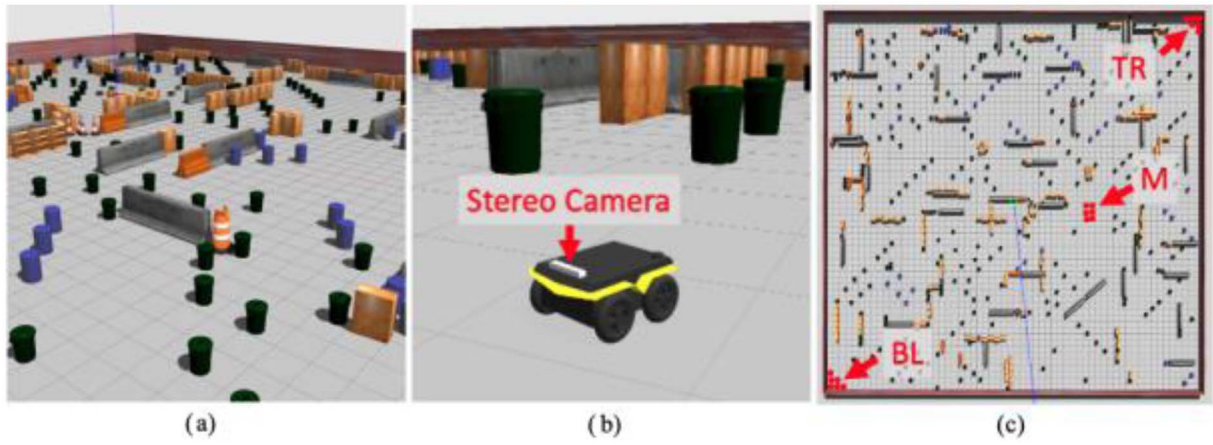

Fig. 6. Multi-robot exploration simulation environment, (a) cluttered, unstructured environment with varying obstacles, b) wheeled differential drive robot with stereo camera, (c) bird's eye view of a $60 \times 60$ $\mathrm{m}^2$ environment set-up, and the robot team initial positions, top right (TR), middle (M), and bottom left (BL).

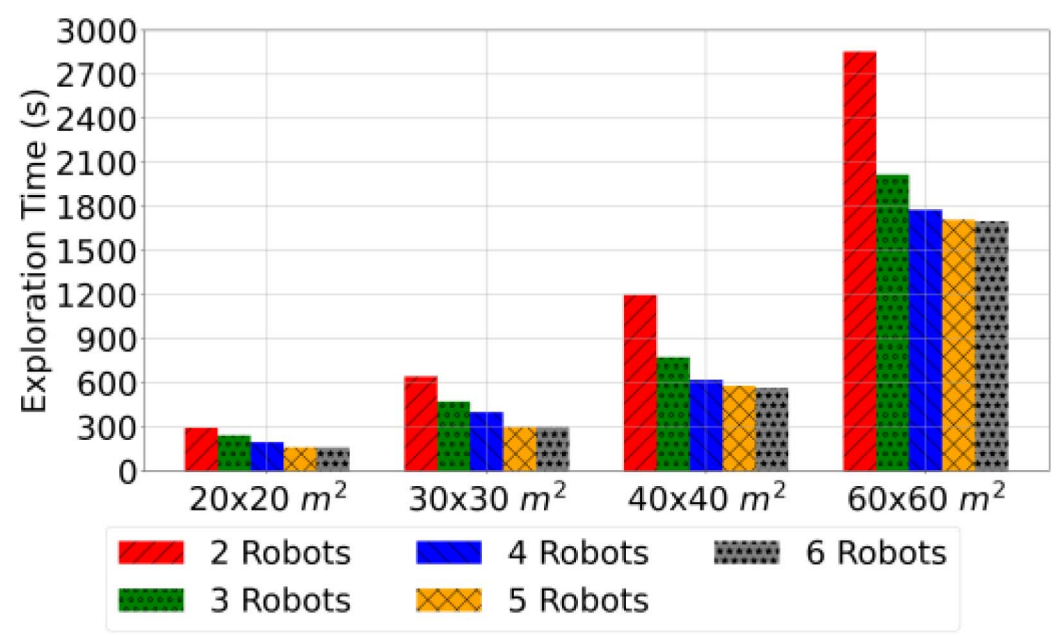

Fig. 7. Average exploration time for each environment and team size.

### B. Scalability Study

We evaluated MADE-Net's performance in terms of exploration time (ET), for increasing environment and team sizes. The experiments were conducted in realistic 3D simulated cluttered, unstructured, environments in Gazebo using the Robot Operating System (ROS) framework.

*1) Environment and Mobile Robot Setup:* Four environments with sizes of $20 \times 20$ $\mathrm{m}^2$, $30 \times 30$ $\mathrm{m}^2$, $40 \times 40$ $\mathrm{m}^2$, and $60 \times 60$ $\mathrm{m}^2$ were randomly generated with obstacles including bookshelves, garbage bins, and cylindrical objects to create an unstructured environment with narrow passageways, dead-ends, and sharp turns, Fig. 6(a). The goal was to investigate MADE-Net's ability to maintain spatial distribution in challenging cluttered environments of varying sizes and with different robot team sizes.

Teams of 2 to 6 mobile robots with non-holonomic differential drive systems, Fig. 6(b), were deployed. Onboard sensors included a stereo camera, an inertial measurement unit (IMU) and wheel encoders for mapping, obstacle avoidance and localization. For each environment and team size combination, a separate MADE-Net model was trained to account for differences in the observation and action spaces. As the joint observation and action space increases with environment and robot team size, training convergence scales proportionally. Simultaneous localization and mapping was achieved using Real Time Appearance-based Mapping (RTAB-Map) [39]. The output of RTAB-Map is an occupancy grid map, where each pixel value represents an occupancy probability. The pixel map was discretized into $1 \times 1$ m cells, where a cell is considered occupied when more than 25% of its pixels have an occupancy probability $> 65\%$. The resultant binary occupancy grid map was used by MADE-Net for exploration goal selection. Each robot used a local Timed Elastic Band Planner (TEB) [40] to account for: 1) navigation among dynamic obstacles by predicting nearby teammates' motions with a constant velocity model to avoid collisions, and 2) the non-holonomic constraint by restricting the mobile robot's lateral velocity to zero during path optimization.

*2) Procedure:* Three trials were conducted with initial team locations in bottom left (BL) corner, top right (TR) corner, and middle (M) of environment, for each environment and team size configuration, for a total of 60 trials, e.g., Fig. 6(c).

*3) Results:* Fig. 7 presents the average total ET for each team and environment size. Overall, the total ET decreased with increasing team size as a result of the increase in joint area coverage. However, as team sizes increased, the individual robots spent more time navigating between exploration goals in order to avoid collisions with teammates. Therefore, at some point increasing the robot team size will stop having an effect on reducing the ET. A Friedman Test showed a statistically significant difference in ET for all team sizes across each environment size $(p < 0.05)$. However, post-hoc analysis using Wilcoxon Signed-Rank tests showed no statistical difference $(p > 0.05)$ between team sizes of 4 and 5 robots, 5 and 6 robots, and 4 and 6 robots in $20 \times 20$ $\mathrm{m}^2$, $40 \times 40$ $\mathrm{m}^2$, and $60 \times 60$ $\mathrm{m}^2$ environments. In the $30 \times 30$ $\mathrm{m}^2$ environment, no statistical difference was found between 5 and 6 robots $(p > 0.05)$. In general, MADE-Net was able to maintain spatial distribution with increasing team and environment sizes. A video of our MADE-Net approach in cluttered environments with both varying CSPs and team and environment sizes is presented on our YouTube channel at https://youtu.be/iTzPRoOS3Q0.

## VI. Conclusion

In this letter, we present a novel macro action based DRL planner for decentralized multi-robot exploration in cluttered and unstructured environments with communication dropouts. MADE-Net learns robot coordination through centralized training while using only local perception for decentralized exploration execution. During communication dropouts, teammate intentions are implicitly considered via the hidden states of the decentralized exploration policy as a property of the resultant macro action. Simulated experiments showed our MADE-Net method had a better overall performance with various degrees of communication dropouts when compared to classical and DRL methods. A scalability study demonstrated MADE-Net's ability to scale to both increasing team and environment sizes. Future work includes: 1) extending the proposed MFE to use pixel values from maps in order to account for arbitrary environment sizes, and 2) integrating MADE-Net with a physical robot team for testing in real-world environments.